\documentclass[letterpaper]{article} 
\usepackage{aaai2026}  
\usepackage{times}  
\usepackage{helvet}  
\usepackage{courier}  
\usepackage[hyphens]{url}  
\usepackage{graphicx} 
\usepackage{graphicx} 
\usepackage{natbib}  
\usepackage{caption} 
\usepackage{algorithm}
\usepackage{algorithmic}

\usepackage{newfloat}
\usepackage{listings}
\DeclareCaptionStyle{ruled}{labelfont=normalfont,labelsep=colon,strut=off} 
\floatstyle{ruled}
\newfloat{listing}{tb}{lst}{}
\floatname{listing}{Listing}
\usepackage[utf8]{inputenc} 
\usepackage[T1]{fontenc}    
\usepackage{booktabs}       
\usepackage{amsfonts}       
\usepackage{nicefrac}       
\usepackage{microtype}      
\usepackage{xcolor}         
\usepackage{amsmath}
\usepackage[utf8]{inputenc} 
\usepackage{url}            
\usepackage{booktabs}       
\usepackage{amsfonts}       
\usepackage{nicefrac}       
\usepackage{microtype}      
\usepackage{subfig}
\usepackage{makecell}
\usepackage{multirow}
\usepackage{bbding}
\usepackage[table]{xcolor}
\usepackage{natbib}
\usepackage{pifont}
\usepackage{xspace}
\usepackage[capitalize]{cleveref}

\crefname{section}{Sec.}{Secs.}
\Crefname{section}{Section}{Sections}
\Crefname{table}{Table}{Tables}
\crefname{table}{Tab.}{Tabs.}

\newcommand{\methodname}{\textit{FFSE}\xspace}
\newcommand{\methodcompletename}{\textit{Free-Form Scene Editor}\xspace}
\newcommand{\normalmethodcompletename}{Free-Form Scene Editor\xspace}
\newcommand{\datasetname}{\textit{3DObjectEditor}\xspace}

\newcommand{\domainadaptercompletename}{\textit{Domain LoRA}\xspace}

\newcommand{\operationselfattentioncompeletename}{\textit{operation self-attention}}
\newcommand{\objectenhancercompletename}{\textit{context self-attention}\xspace}
\newcommand{\objectenhancername}{\textit{CSA}\xspace}

\newcommand{\frameencodercompletename}{\textit{frame encoder}\xspace}

\newcommand{\operationencodercompletename}{\textit{operation encoder}\xspace}

\newcommand{\xmarkg}{\textcolor{lightgray}{\ding{55}}\xspace}%
\newcommand{\cmark}{\ding{51}\xspace}%

\definecolor{myorange}{RGB}{255,100,3}
\definecolor{mygray}{gray}{.85}
\definecolor{mygray1}{gray}{.7}
\definecolor{mygray2}{gray}{.93}
\definecolor{mygray3}{gray}{.90}

\newcommand{\ie}{{\emph{i.e.}}\xspace}
\newcommand{\eg}{{\emph{e.g.}}\xspace}

\title{Free-Form Scene Editor: Enabling Multi-Round Object Manipulation\\ like in a 3D Engine}
\author{
    Xincheng Shuai\textsuperscript{\rm 1},
    Zhenyuan Qin\textsuperscript{\rm 1},
    Henghui Ding\textsuperscript{\rm 1}\thanks{Corresponding author.},
    Dacheng Tao\textsuperscript{\rm 2} \\
}
\affiliations{
    \textsuperscript{\rm 1}Fudan University, \textsuperscript{\rm 2}Nanyang Technological University\\
    henghui.ding@gmail.com, 
    dacheng.tao@gmail.com\\
    \url{https://henghuiding.com/FFSE/}
}

\usepackage{bibentry}

\begin{document}

\maketitle

\begin{abstract}
Recent advances in text-to-image (T2I) diffusion models have significantly improved semantic image editing, yet most methods fall short in performing 3D-aware object manipulation. In this work, we present \methodname, a 3D-aware autoregressive framework designed to enable intuitive, physically-consistent object editing directly on real-world images. Unlike previous approaches that either operate in image space or require slow and error-prone 3D reconstruction, \methodname models editing as a sequence of learned 3D transformations, allowing users to perform arbitrary manipulations, such as translation, scaling, and rotation, while preserving realistic background effects (e.g., shadows, reflections) and maintaining global scene consistency across multiple editing rounds. To support learning of multi-round 3D-aware object manipulation, we introduce \datasetname, a hybrid dataset constructed from simulated editing sequences across diverse objects and scenes, enabling effective training under multi-round and dynamic conditions. Extensive experiments show that the proposed \methodname significantly outperforms existing methods in both single-round and multi-round 3D-aware editing scenarios.
\end{abstract}

\begin{links}
\link{Code}{https://github.com/FudanCVL/FFSE}
\end{links}

\section{Introduction}

\begin{figure*}[t]
    \centering
    \includegraphics[width=1\linewidth]{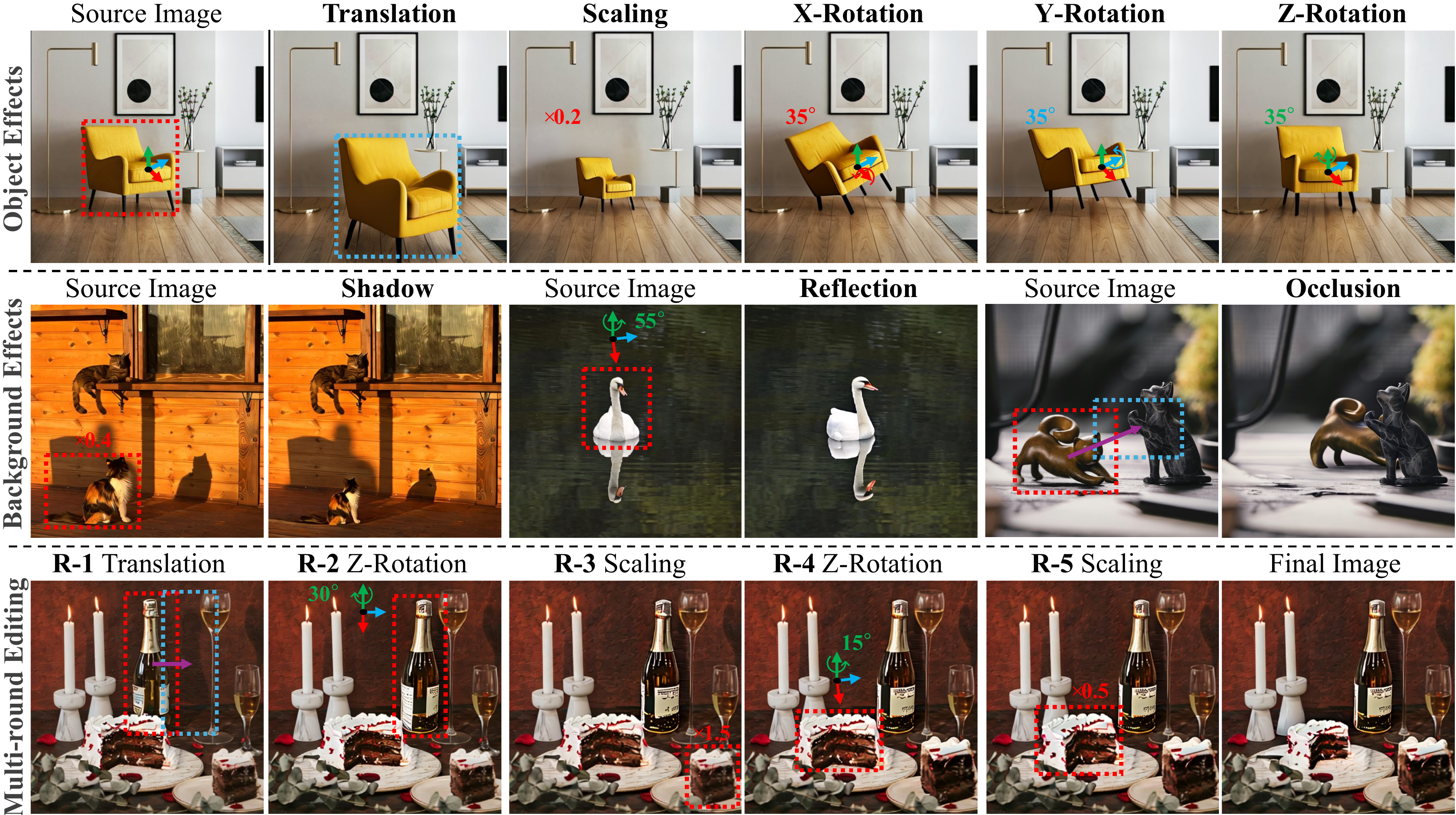}
    \caption{3D-aware object manipulation results of our \methodcompletename (\methodname).
    {1)~Object effects.}~\methodname can process a variety of 3D operations, including challenging transformations such as rotations.~{2)~Background effects.}~\methodname generates realistic environmental interaction resulting from object manipulations, such as shadows and occlusions.~{3)~Multi-round editing.}~\methodname maintains consistency of scene elements across multiple editing iterations. {Moreover, the proposed \methodname provides a user-friendly interface without time-consuming 3D reconstruction.}}
    \label{fig:teaser}
\end{figure*}

Image editing enables users to modify the visual content without requiring expertise in professional software. Recently, advanced methods~\cite{DragDiffusion,Editable-Image-Elements} based on text-to-image (T2I) diffusion models \cite{StableDiffusion,StableDiffusionXL} enable powerful object-centric manipulation, including appearance~\cite{P2P,PnP,InstructPix2Pix} or shape modification~\cite{DragDiffusion,FreeDrag}. Although these methods are useful for semantic editing, 3D-aware object manipulation~\cite{SceneDesigner} offers more flexibility in many scenarios, as shown in \cref{fig:teaser}. By equipping models with 3D-aware editing ability, users can manipulate objects as if operating within a 3D engine~\cite{unrealengine}, allowing more intuitive and physically-consistent image modifications.

Recently, a growing body of works~\cite{Object-3DIT,Image-Sculpting,Diff3DEdit,DiffusionHandles,3DitScene,DORSal,NeuralAssets} explore 3D-aware image editing, but still struggle to produce satisfactory results. \textit{Image-space} methods~\cite{Zero-1-to-3,NeuralAssets,Object-3DIT} typically learned 3D priors from established datasets. However, most of methods fail to accomplish comprehensive 3D operations and demonstrate poor generalization on real-world images. \textit{3D-space} methods~\cite{DiffusionHandles,Image-Sculpting,OMG3D,Diff3DEdit} reconstruct scene structure from a single image to support arbitrary 3D manipulations, but are time-consuming and sensitive to noisy geometry estimations~\cite{StableDiffusion,SDS,NeRF,Zero-1-to-3}. Moreover, most existing methods struggle to generate realistic background effects (\eg, shadows, reflections) and often fail to maintain scene consistency across multi-round edits.

Several key challenges remain in achieving effective 3D-aware object manipulation: \textbf{1) Object effects.} Existing methods~\cite{Object-3DIT,Zero-1-to-3} support only limited and simple 3D operations, or often yield low-quality results in complex scenarios~\cite{3DitScene,Image-Sculpting,DiffusionHandles}. \textbf{2) Background effects.} Current approaches struggle to infer realistic environmental interactions caused by object manipulations, such as shadows, reflections, and occlusions. \textbf{3) Multi-round editing.} Existing studies lack awareness of scene structure changes, resulting in inconsistencies after multiple edits. \textbf{4) User interface.} Previous works often rely on cumbersome inputs~\cite{NeuralAssets,Object-3DIT}, or require time-consuming reconstruction process~\cite{Image-Sculpting,3DitScene}.

To address the above challenges, we propose \methodcompletename\ (\methodname), an autoregressive generation model that is capable of handling diverse 3D operations in real-world images, akin to modern 3D engines such as \textit{Unreal}~\cite{unrealengine} and \textit{Blender}~\cite{blender}. A straightforward way is to leverage existing video datasets~\cite{WebVid-10M,RealEstate-10K} and estimate corresponding transformations using off-the-shelf tools~\cite{RAFT,sam,Magic-Fixup,NeuralAssets}. However, such data often contains undesired background dynamics and noisy annotations, making it unsuitable for learning precise object-centric manipulations. To address this issue, we construct a hybrid dataset, \datasetname, which employs image sequences generated with physical simulation to emulate realistic manipulation dynamics. Unlike existing datasets~\cite{Object-3DIT}, \datasetname is specifically designed to support multi-round editing and covers a broader range of object and scene categories, along with higher domain diversity. Trained on \datasetname, our \methodname model can perform various 3D operations iteratively, while generating plausible background effects and maintaining scene consistency across edits. In addition, our multi-stage training strategy enables the model's robust performance on real-world images.

In summary, our main contributions are as follows:
\textbf{1)} We introduce \datasetname, a hybrid dataset that simulates image sequences resulting from diverse 3D operations across a wide range of objects and scenes. 
\textbf{2)} We propose \methodname, an autoregressive framework for 3D-aware image editing that produces realistic object and background effects while maintaining scene consistency.
\textbf{3)} Extensive experiments demonstrate the superior performance of the proposed \methodname over existing methods.

\section{Related Works}
\noindent\textbf{Image Editing.}
Image editing~\cite{SDEdit,P2P,shuai2024survey,MagicRemover} has been significantly enhanced by powerful text-to-image (T2I) diffusion models. Existing models enable fine-grained image editing, like appearance modification, object removal, and image inpainting.

\noindent\textbf{Object-centric Spatial Manipulation.}
Current methods for object-centric manipulation rely on spatial cues like point pairs~\cite{FreeDrag,DragDiffusion} or 2D masks~\cite{DragonDiffusion,Diffusion-Self-Guidance,FFMC,AnyI2V}, but struggle with precise object-level transformations and 3D-aware operations. While some approaches reconstruct 3D scenes for better control~\cite{OMG3D,Image-Sculpting}, they suffer from time-consuming optimization and noisy geometry estimation.

\section{Methodology}\label{sec:methodology}
\subsection{Overview}

To simulate the behavior of interactive 3D engines~\cite{blender,unrealengine} and enable manipulation of real-world images, our goal is to train a neural network for iterative 3D-aware object manipulation. We first define a formal setting involving a scene state space $S$, an operation space $O$, and a state transition function $p_{tf}(s^{\prime}|s,o)$, {where $s, s^{\prime} \in S$ represent the current and next scene states, and $o \in O$ denotes an editing operation}. In addition, we define an observation space $X$, where each element is a projected view of the state space $S$ via a mapping function $f_m$.

Specifically, {the state space $S$ represents the underlying configuration of scene elements, such as object and background components, while the observation space $X$ corresponds to images captured from camera views through the rendering function $f_m$}. The operation space $O$ includes a set of 3D transformations $\left\{o^{T}, o^{S}, o^{X}, o^{Y}, o^{Z}\right\}$, corresponding to translation, scaling, and rotations around $x$/$y$/$z$ axes aligned with the forward, left, and upward directions in the object space, respectively. The transition function $p_{tf}(s^{\prime}|s,o)$ can be implemented by a physical simulator that updates the scene state based on the current state and applied operation. 

Given the editing history $h_{r}=\left\{(x_i,o_i)\right\}_{i=0}^{r-1}$ up to $r$-th round, where $x_i \in X$ and $x_0$ is the source image, our goal is to model the observation distribution $p(x_r|h_r)$ using a diffusion-based generative model $p_{\theta}$. To train this model, we collect a dataset $\{(h_{r_j}, x_{r_j})\}_{j=1}^{N_D}$ via a rule-based data generation pipeline below.

\subsection{Data Generation}\label{sec:data_generation}

There is no well-established public dataset tailored for learning 3D-aware object manipulation in multi-round editing. A suitable dataset must meet several key criteria (see appendix), while the existing data sources violate some of them. Therefore, we introduce \datasetname, a hybrid dataset that combines real $D_\text{real}$ and synthetic $D_\text{syn}$ samples.

\noindent\textbf{Realistic Domain $D_\text{real}$.} We construct image sequences in $D_\text{real}$ using following steps: \textbf{1) Asset collection.} We leverage MULAN~\cite{MULAN}, which contains RGBA images decomposed from scenes in MS-COCO~\cite{COCO} and LAION~\cite{LAION-400M} datasets. We only use the MS-COCO part.~\textbf{2) Scene construction.} Based on labeled foreground \& background tags, we randomly select a background and several object images as scene elements. For each object, its center position ${(x_p,y_p)}$ in the background image and depth value $d$ are initialized randomly.~\textbf{3) Sequence construction.} For each editing step, an object is randomly chosen for manipulation, with the operation sampled from the restricted space $\left\{o^T,o^S\right\}$ in $D_\text{real}$. To simulate different distances from camera view, we determine the object's size (the shortest side of image) through $(d-d_\text{max})\frac{s_\text{min}-s_\text{max}}{d_\text{max}-d_\text{min}}+s_\text{min}$, where $[s_\text{min},s_\text{max}]$ and $[d_\text{min},d_\text{max}]$ denote size and depth bounds, respectively. The size bounds $s_\text{min}$ and $s_\text{max}$ are computed by scaling their predefined values $\hat{s}_\text{min}$ and $\hat{s}_\text{max}$ with the object’s current scaling factor $f_s$, \ie, $s = \hat{s} \times f_s$. During translation, we randomly update the object’s position $(x_p, y_p)$ and depth $d$, while in scaling, $f_s$ is adjusted. The final image is rendered using the painter’s algorithm: the background is placed first, followed by foreground objects rendered in depth order from farthest to nearest.

\noindent\textbf{Synthetic Domain $D_\text{syn}$.} Relying solely on $D_\text{real}$ is insufficient, as it lacks support for precise object rotation and realistic background effects. Therefore, we employ Blender~\cite{blender} for high-fidelity simulation: \textbf{1) Asset collection.} High-resolution panoramic images and 3D scenes from PolyHaven~\cite{PolyHaven} and Sketchfab~\cite{Sketchfab} are used as backgrounds. For objects, over 6,000 assets are filtered from Objaverse-LVIS~\cite{objaverse} and Objaverse-XL~\cite{objaverse-xl}, including animated models to increase pose diversity.  \textbf{2) Scene construction.} A background and a set of object assets are randomly selected to form a scene. Directional and point lights are added to avoid underexposed renderings. Objects are appropriately scaled and randomly placed on the ground plane. For animated assets, a random keyframe is selected to provide diverse object poses. \textbf{3) Sequence construction.} At each step, any 3D operation from $\left\{o^{T}, o^{S}, o^{X}, o^{Y}, o^{Z}\right\}$ is applied to a randomly chosen object. Finally,  Blender’s Cycles ray tracer simulates realistic physics, producing high-quality environmental interaction such as shadows.

\subsection{\normalmethodcompletename}\label{sec:framework}
Existing methods~\cite{Object-3DIT,Image-Sculpting,Diff3DEdit,DiffusionHandles,3DitScene,NeuralAssets} demonstrate limited capability in 3D-aware object manipulation, particularly in handling realistic \textit{object effects}, \textit{background effects}, and maintaining consistency in \textit{multi-round editing}. To address these challenges, we propose \methodcompletename(\methodname), an approach that approximates $p(x_r | h_r)$ using a neural network parameterized by $\theta$. The model estimates the new observation $x_r$ in an autoregressive manner, conditioned on the editing history $h_r$.

\begin{figure*}[t]
    \centering
    \includegraphics[width=0.999\linewidth]{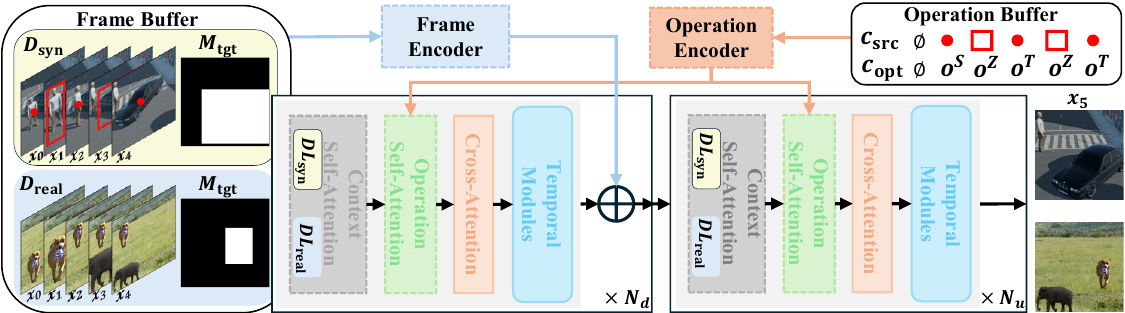}
    \caption{Overall framework of \methodcompletename (\methodname) with dashed boxes indicating introduced learnable modules, where the middle blocks and convolutional layers from the base model are omitted for simplicity. $N_d$ and $N_u$ denote the number of down and up blocks, respectively. Two 6-length editing sequences are shown as an example, where only $DL_\text{syn}$ is active since the current training batch is sampled from $D_\text{syn}$. Historical observations and operations are processed by \frameencodercompletename and \operationencodercompletename, respectively, to capture scene structure changes. The output of \frameencodercompletename is added to down block features, while the output from \operationencodercompletename is injected into the main branch via \operationselfattentioncompeletename. Additionally, standard self-attention modules are enhanced by \objectenhancercompletename to improve the appearance consistency of the edited object.}
    \label{fig:pipeline}
\end{figure*}

The overall framework of \methodname is shown in \cref{fig:pipeline}. For training efficiency, we build upon a pretrained video generation backbone~\cite{SVD}. To model the sequential editing process, we maintain a frame buffer and an operation buffer to store historical observations and user-specified operations.
To encode $h_r$, \operationencodercompletename processes the past operations $\left\{o_i\right\}_{i=0}^{r-1}$ from the operation buffer. The outputs are injected into the main branch via \operationselfattentioncompeletename~to guide the editing behavior. In parallel, \frameencodercompletename encodes previous observations $\left\{x_i\right\}_{i=0}^{r-1}$ to capture scene dynamics and structural context. It also receives a target location mask to guide object placement in the current step. To improve the consistency of the edited object, we introduce \objectenhancercompletename to enhance the standard self-attention layers. Furthermore, to prevent learning the domain-specific content, we introduce \domainadaptercompletename for each domain in \datasetname, and employ a multi-stage training strategy to ensure robust performance on real-world images.

\noindent \textbf{Encoding of the History.} We apply different components to process operations and observations, respectively. Specifically, the operation $o_i$ in round $i$ can be decomposed into \textit{source region} and \textit{operation type\&value}. In our implementation, the centroid $l_i^{p}$ and bounding box $l_i^{b}$ are used as \textit{source region} to locate the object before manipulation with different granularities. On the other hand, \textit{operation type\&value} indicates the relative transformation of the object from the last frame. Concretely, $o_i^{T}$ is presented by the normalized pixel offset, while $o_i^{S}$ and $o_i^{X},o_i^{Y},o_i^{Z}$ are encoded by the scaling factor and rotation angle around the corresponding axis. Formally, \operationencodercompletename encodes \textit{source region} and \textit{operation type \& value} by fourier embedding and MLP, which is denoted by $f(\cdot)=\text{MLP}(\text{Fourier}(\cdot))$. The encoded features of \textit{source region} $c_i^\text{src}$ and \textit{operation type \& value} $c_i^\text{opt}$ in the $i$-th round are formalized as:
\begin{equation}
\begin{aligned}\label{eq:encode}
c_i^\text{src}=&[f(l_i^{p}),f(l_i^{b})], \\
c_i^\text{opt}=&[f(o_i^{T}),f(o_i^{S}),f(o_i^X),f(o_i^Y),f(o_i^Z)],
\end{aligned}
\end{equation}
where $[\space \cdot \space ]$ is channel-wise concatenation. Besides, we set the corresponding elements to learnable ``null'' embeddings for conditions not provided. As shown in~\cref{fig:pipeline}, we concatenate the "null" conditions $\emptyset$ with $\big\{c_i^\text{src}\big\}_{i=0}^{r-1}$ and $\big\{c_i^\text{opt}\big\}_{i=0}^{r-1}$ along the sequence dimension, denoted as $c_{\text{src}}$ and $c_{\text{opt}}$, respectively, where $\emptyset$ indicates that the initial observation $x_0$ is not derived from any operation. Finally, inspired from previous works~\cite{GLIGEN}, we integrate assembled features $[c_\text{src},c_\text{opt}]$ into the network by injecting \operationselfattentioncompeletename~between the \textit{context self-} and cross-attention layers from spatial modules: 
\begin{equation}
\begin{aligned}\label{eq:gated_self_attention}
\hat{v}=\bar{v}+\beta\cdot \text{tanh}(\gamma)\cdot \text{TS}\big(\text{SelfAttn}([\bar{v},\text{repeat}([c_\text{src},c_\text{opt}])])\big),
\end{aligned}
\end{equation}
where $\bar{v}$ is the features from \objectenhancercompletename. Concretely, $[c_\text{src},c_\text{opt}]$ is repeated in spatial dimension to align with $\bar{v}$. $\text{TS}$ truncates the features to select visual tokens. $\gamma$ is a learnable scalar and $\beta$ modulates the control effect in inference time.

On the other hand, we represent \textit{target region} by the binary mask $M_\text{tgt}$, which is derived from the bounding box of the target location in the current round. This design enhances the location accuracy of the edited object. Then, it is concatenated with previous observations $\left\{x_j\right\}_{j=0}^{r-1}$, and the final input is fed to the \frameencodercompletename as indicated in \cref{fig:pipeline}, which is a lightweight branched network. Finally, the output is added to the spatial features from the down blocks. We randomly omit $M_\text{tgt}$ during training by applying an all-zero mask, which allows the model to implicitly learn the \textit{target region} from the current operation and observations, avoiding cumbersome input from users in inference time.

\noindent\textbf{Context Self-attention.} To maintain the appearance consistency of edited objects after operations, we introduce \objectenhancercompletename (\objectenhancername) to enhance the ordinary self-attention modules by referring the edited object in the $r$-th round to the same object from the last observation, expressed as:
\begin{equation}
\begin{aligned}\label{eq:context_self_attention}
\bar{v}_r=v_r+\lambda M_\text{tgt} \text{softmax}(A_{r,r-1}+\frac{Q_r^{'}(K_{r-1}^{'})^T}{\sqrt{d}})V_{r-1}^{'},
\end{aligned}
\end{equation}
where subscripts $r$ and $r-1$ represent sliced features corresponding to the current and the last rounds, respectively, and $'$ indicates that the variables are calculated by newly injected layers. For example, $v_r$ is the $r$-th round feature from ordinary self-attention modules. The learnable $\lambda$ adjusts the effects from the last observation, and $M_\text{tgt}$ prevents the influence in irrelevant pixels. Furthermore, $A_{r,r-1}$ is a $hw \times hw$ matrix and the element in $[i,j]$ is set to $0$ only if $\text{Vec}(M_{r})[i]$ and $\text{Vec}(M_{r-1})[j]$ are all located in the object region, where $\text{Vec}(\cdot)$ represents that the matrix is flattened to vector, $M_r$ and $M_{r-1}$ are masks derived by object bounding boxes in the current and the last steps. Other elements in $A_{r,r-1}$ are all set to an infinitesimal value, ensuring that only the features inside the object region are involved in mutual computation.

\begin{figure*}[t]
    \centering
    \includegraphics[width=0.999\linewidth]{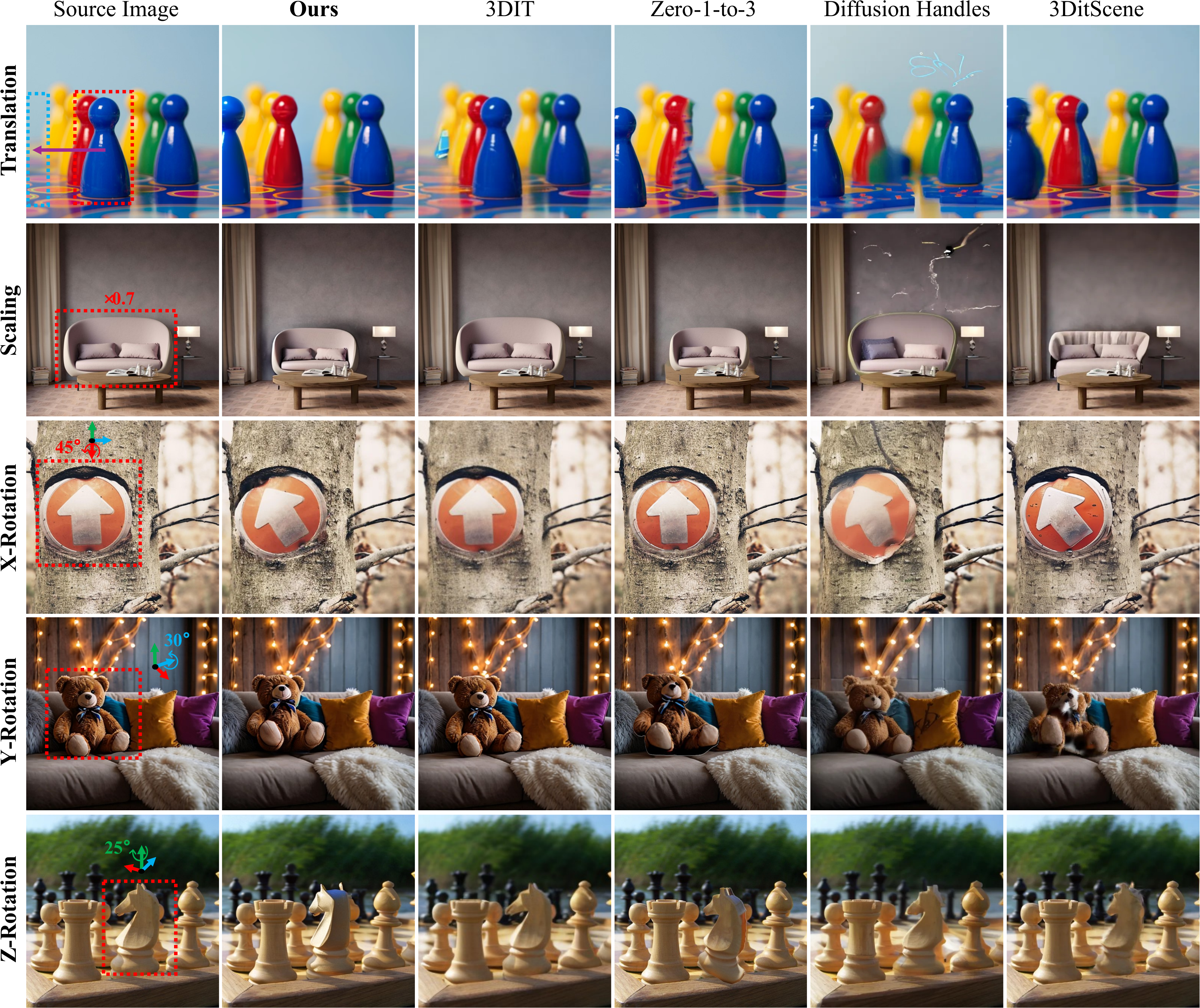}
    \caption{Evaluation of object effects under different 3D operations in single-round editing.}
    \label{fig:exp1}
\end{figure*}

\noindent \textbf{Multi-stage Training Strategy.} We propose a multi-stage training strategy to learn from \datasetname. First, we train the newly injected modules with the whole dataset. To prevent overfitting to domain-specific content from different sources, we adopt \domainadaptercompletename~\cite{LoRA} $DL_\text{real}$ and $DL_\text{syn}$ for $D_\text{real}$ and $D_\text{syn}$, respectively. During training, the corresponding LoRA modules are loaded into the network according to the domain identifier of the sample. This setup allows the model to capture object effects caused by specific operations shared between $D_\text{real}$ and $D_\text{syn}$, while enhancing generalization for diverse in-the-wild images. As shown in \cref{fig:pipeline}, these LoRA modules are injected only into the \objectenhancercompletename layers. The training objective for this stage is:
\begin{equation}
\begin{aligned}\label{eq:stage1_loss}
\underset{\theta,DL_\text{real,syn}}{\arg \min } E_{(h_{r},x_r)\sim D_\text{real,syn}}\left[\left\|\varepsilon_{\theta,DL_\text{real,syn}}\left(x_{0:r}^t, t, h_{r}\right)-\epsilon\right\|^2 \right]
\end{aligned}
\end{equation}
where $x_{0:r}^t$ is the noisy image sequence in time $t$, obtained by injecting Gaussian noise $\epsilon$ into $x_{0:r}$ via diffusion forward process. $r$ is sampled from a uniform distribution bounded by $[r_\text{min},r_\text{max}]$.

Due to the insufficient simulation in $D_\text{real}$, the model trained with \cref{eq:stage1_loss} may generate unrealistic results. To enhance the quality of generated background effects such as shadows, we finetune the model solely with $D_\text{syn}$. In this stage, we load $DL_\text{syn}$ into network while only optimizing $\theta$:
\begin{equation}
\begin{aligned}\label{eq:stage2_loss}
\underset{\theta}{\arg \min } E_{(h_{r},x_r)\sim D_\text{syn}}\left[\left\|\varepsilon_{\theta,DL_\text{syn}}\left(x_{0:r}^t, t, h_r\right)-\epsilon\right\|^2 \right].
\end{aligned}
\end{equation}

\noindent \textbf{Inference.} Our framework supports iterative user interaction, enabling the multi-round object manipulation from an initial source image $x_0$. It is worth noting that \domainadaptercompletename modules are omitted during inference to preserve the quality of the base generative model.

\section{Experiment}\label{sec:experiment}
\textbf{Implementation Details.} We build \methodname upon the pretrained image-to-video model SVD~\cite{SVD}, training it with the Adam optimizer (initial learning rate: $1e^{-4}$) on 4 NVIDIA A800 80G GPUs. All experiments use $512\times512$ resolution and a batch size of 8, with $r_\text{min}=1$ and $r_\text{max}=12$. In Stage 1, we jointly train $\{DL_\text{real}, DL_\text{syn}\}$ and newly introduced parameters $\theta$ using $D_\text{real}$ and $D_\text{syn}$ for 80K iterations. In Stage 2, we load $DL_\text{syn}$ to the model and further optimize $\theta$ on $D_\text{syn}$ for 10K iterations.

\begin{table*}[t]
\centering

\setlength\tabcolsep{6pt}
  \begin{tabular}{c|c|cc|cc}
    \specialrule{0.7pt}{0pt}{0pt}
    \multirow{2}{*}{\textbf{Task}} & \multirow{2}{*}{\textbf{Method}}   & \multicolumn{2}{c|}{\textbf{Content Consistency}} & \multicolumn{2}{c}{\textbf{Identity Preservation}}  \\
    \cline{3-6}
    
    & & PSNR $\uparrow$ & SSIM$_{\times 10^{2}}$ $\uparrow$ & DINO $\uparrow$ & CLIP $\uparrow$  \\ \hline
\multirow{5}{*}{Single-round Editing}  &3DIT~\cite{Object-3DIT}&20.12&68.76&61.38&80.96\\
 &Zero-1-to-3~\cite{Zero-1-to-3} &\underline{23.84}&\underline{71.97}&65.42&83.27\\
& Diffusion Handles~\cite{DiffusionHandles} &18.83&58.33&71.33&88.53 \\
 & 3DitScene~\cite{3DitScene} &17.67&53.39&\underline{73.69}&\underline{89.11}\\
 &\textbf{\methodname} (\textbf{ours}) &\textbf{26.31}&\textbf{79.54}&\textbf{82.39}&\textbf{91.67}\\

\hline
 \multirow{5}{*}{Multi-round Editing} &3DIT&18.31&57.62&60.19&78.27 \\
 &Zero-1-to-3 &\underline{19.81}&\underline{64.77}&\underline{61.67}&\underline{82.38}\\
 &Diffusion Handles&13.79&50.47&59.06&78.24\\
 &3DitScene&10.75&43.24&42.17&76.35\\
 &\textbf{\methodname} (\textbf{ours}) &\textbf{24.96}&\textbf{74.99}&\textbf{79.51}&\textbf{90.42}\\
\hline
\specialrule{0.7pt}{0pt}{0pt}
  \end{tabular}
\caption{Quantitative evaluation in single-round and multi-round editing.}\label{tab:quantitative}
\end{table*}

\noindent
\textbf{Evaluation Details.}  For assessing the performance of models, we concentrate on the following aspects.~\textbf{1) Maintenance of source content.} We use PSNR and SSIM for assessing the overall visual similarity. ~\textbf{2) Identity preservation of the edited object.} For the operated object, we use CLIP-Score~\cite{CLIP} and DINO-Score~\cite{DINO} to compute the semantic similarity between the objects before and after the manipulation. For multi-round editing, we average the results of adjacent image pairs from generated frames. Furthermore, since the operation effects on the object and background are difficult to evaluate, we conduct a user study to represent the human preference. For the validation dataset, we source high-quality images from public websites like Unsplash~\cite{Unsplash}, Pixabay~\cite{pixabay}, and Pexels~\cite{Pexels} to construct diverse and complex scenes from the real world. Specifically, we collect 50 images in total and manually select the objects to be edited. Next, we use an off-the-shelf tool~\cite{G-DINO} to estimate the bounding box and centroid as \textit{source region}. Then, several \textit{target region} and \textit{operation type\&value} pairs are assigned for objects to construct samples for different operations. We also construct operation sequences to assess multi-round editing performance. Finally, for single-round editing, we obtain 45 test cases for translation, 48 for scaling, 30 for rotation around the $x/y$ axis, and 40 for rotation around the $z$ axis. We also get 30 examples for the multi-round editing experiment, where the sequence length is fixed to 6.

\noindent
\textbf{Compared Methods.} For compared methods, we focus on two algorithm families and choose the methods that are open source and can handle most of the 3D operations described in our paper. For image space methods, we choose 3DIT~\cite{Object-3DIT} and Zero-1-to-3~\cite{Zero-1-to-3} as compared methods. Specifically, we crop the source region from the image and apply Zero-1-to-3 to get the transformed object, which is then overlaid on the target region of the inpainted background image. Compared 3D methods include Diffusion Handles~\cite{DiffusionHandles} and 3DitScene~\cite{3DitScene}, which manipulate estimated point clouds or reconstructed 3DGS~\cite{3DGS}. Since 3DitScene and Diffusion Handles do not implement scaling operations, we move the object from/closer to the camera to emulate scaling down/up. Besides, for operations that are not supported by the compared method, we directly return the source image.

\subsection{Comparisons with State-of-the-Art Methods} 
\noindent
\textbf{Comparison on Single-round Editing.} In this experiment, we primarily evaluate the performance in single-round editing, concentrating on the effects of object and background caused by specific manipulations. \cref{fig:exp1} demonstrates the performance of object effects from different 3D operations. For image space methods, 3DIT only supports limited operation types and suffers from poor generalization ability. As a result, it fails to accomplish most of the edits. On the other hand, although it suffers from a cumbersome workflow, Zero-1-to-3 can handle most of the 3D operations. However, it fails to achieve the goal in some complicated scenarios (rows 3,5 in \cref{fig:exp1}). Furthermore, the noisy estimation of the inpainting model leads to unwanted artifacts in the occluded area, as shown in row 1 of \cref{fig:exp1}. In contrast, 3D space methods outperform in operations requiring geometric knowledge, such as object rotation. Nevertheless, they are limited by the time-consuming reconstruction process and low-quality results caused by noisy geometry estimation. In comparison, our method accomplishes all operations with high fidelity and quality. For example, \methodname can recover or manipulate the occluded object (the red checker piece in row 1 and the sofa in row 2). For rotation around principal axes, our method generates accurate object transformations, while achieving realistic physical effects, such as the reflection of light on the chess piece in the last row of \cref{fig:exp1}. The quantitative results in \cref{tab:quantitative} and the user study in the appendix also demonstrate that \methodname outperforms in consistency, quality, and operation fidelity.  

\begin{figure}[t]
    \centering
    \includegraphics[width=0.999\linewidth]{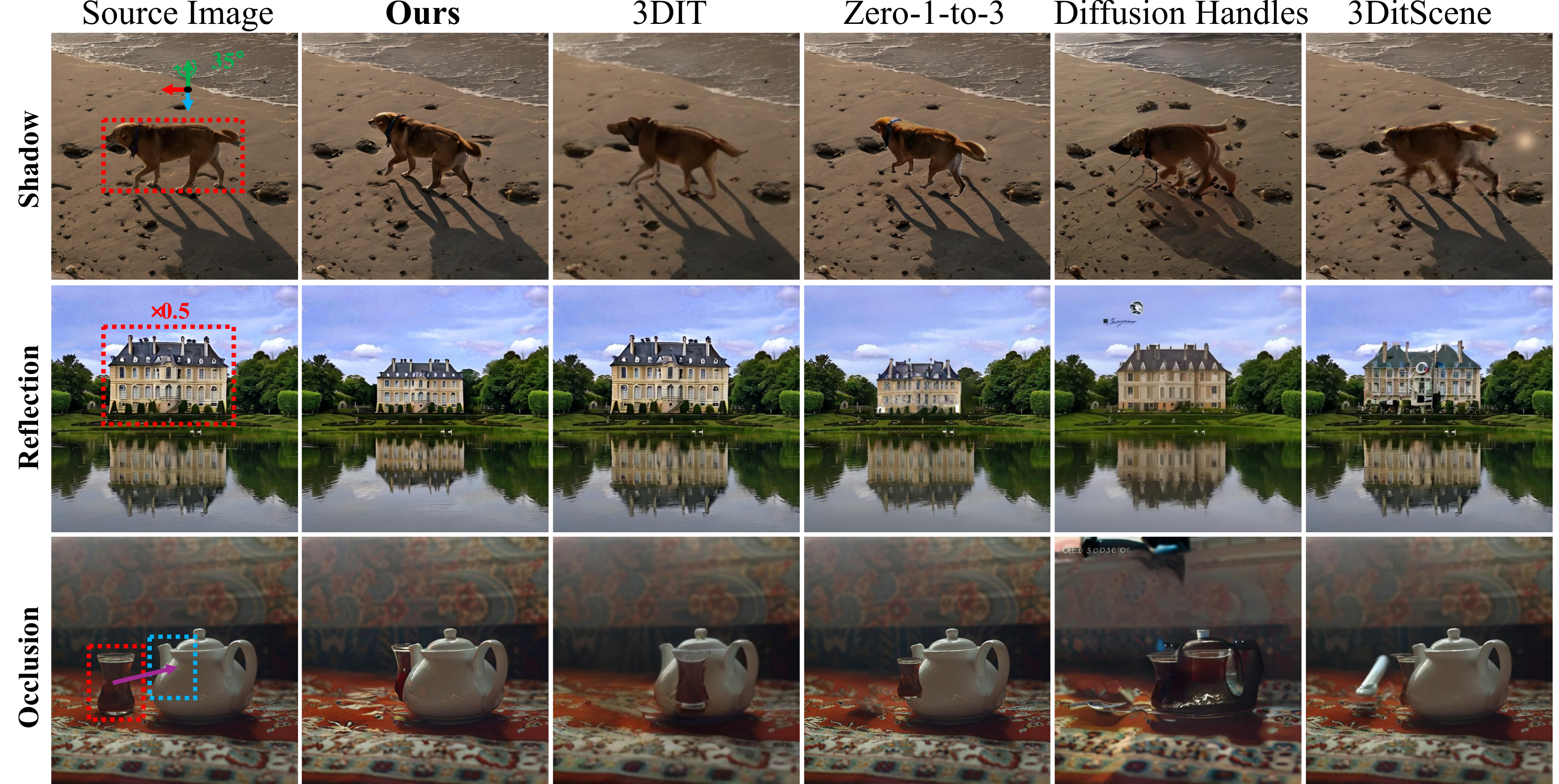}
    \caption{Evaluation of background effects in single-round editing. The figures demonstrate that \methodname generates more physically-plausible environmental interactions.}
    \label{fig:exp2}
\end{figure}

We assess the plausibility of background effects in \cref{fig:exp2}. As shown in rows 1,2 of \cref{fig:exp2}, the compared methods fail to create realistic shadows and reflections, caused by the limited knowledge of the interaction between the object and environment. In the 1st row of \cref{fig:exp2}, most of the methods fail to remove the dog's shadow in the source area and generate the correct effect in the target location. Although 3DIT is trained on data created by physical simulation, it is limited by the poor generalization ability on real-world images. The last row of \cref{fig:exp2} evaluates the case of occlusion. Among the compared methods, 3DitScene achieves better performance using reconstructed scene structure, but with artifacts in the source region. Other methods struggle to place the object behind the teapot, exhibiting incorrect occlusion. Our method generates reasonable environmental interactions in all cases. 
The user study in the appendix also demonstrates the physical plausibility of our method.

\begin{table}[htbp]
\centering

\setlength\tabcolsep{0.3mm}
  \begin{tabular}{c|cc|cc}
    \specialrule{0.7pt}{0pt}{0pt}
    \multirow{2}{*}{\textbf{Setting}}   & \multicolumn{2}{c|}{\textbf{Content Consistency}} & \multicolumn{2}{c}{\textbf{Identity Preservation}}  \\
    \cline{2-5}
    & PSNR $\uparrow$ & SSIM$_{\times 10^{2}}$ $\uparrow$ & DINO $\uparrow$ & CLIP $\uparrow$  \\ \hline
w/ $D_\text{real}$ &25.86&79.31&\underline{81.92}&\underline{91.11} \\
w/ $D_\text{syn}$ &24.37&74.51&73.31&86.43\\
w/o stage 2 &\underline{25.92}&\underline{79.33}&78.77&89.82\\

w/o $DL$ (a) & 25.37&76.54&79.53&89.75\\
w/o $DL$ (b) &24.53&73.25&74.92&88.13\\
 w/o \objectenhancername &24.81&75.17&75.65&88.71\\
\textbf{\methodname}(\textbf{Ours})&\textbf{26.31}&\textbf{79.54}&\textbf{82.39}&\textbf{91.67}\\

\hline
\specialrule{0.7pt}{0pt}{0pt}
  \end{tabular}
\caption{Ablation studies.}\label{tab:ablation_study}
\end{table}

\begin{figure}[t]
    \centering
    \includegraphics[width=0.47\textwidth]{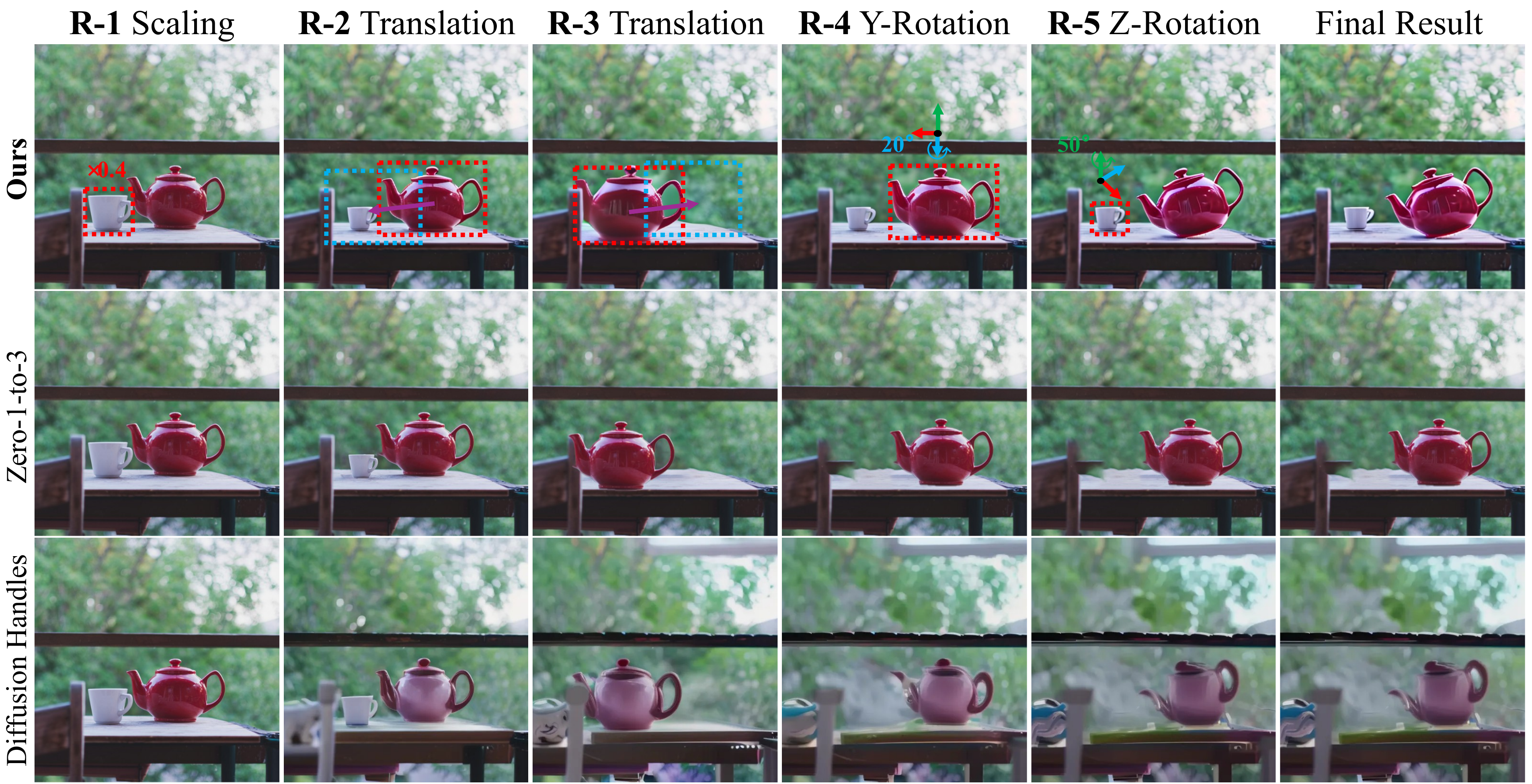}
    \caption{Evaluation in multi-round editing. \methodname accomplishes the operation in each editing round, and maintains high consistency of scene elements. As indicated in columns 2-4 of our method, the cup is first occluded by the teapot, and then becomes visible.}
    \label{fig:exp3}
\end{figure}

\noindent
\textbf{Comparison on Multi-round Editing.} In this experiment, we evaluate the multi-round editing performance of \methodname and representative methods from image space and 3D space algorithm families. For the qualitative experiment, we assess methods' awareness of scene structure changes through changing the occlusion relationships among objects. As shown in columns 2-4 of \cref{fig:exp3}, we translate the teapot in front of the cup and then move it away. Both Zero-1-to-3~\cite{Zero-1-to-3} and Diffusion Handles~\cite{DiffusionHandles} fail to recover the cup. Besides, as the number of editing rounds increases, the quality of their results deteriorates progressively due to accumulated errors. In contrast, our method accurately restores the occluded object, exhibiting high consistency after multiple manipulations. \cref{tab:quantitative} also reveals that \methodname achieves better performance in multi-round editing.

\subsection{Ablation Studies} 
\begin{figure}[t]
    \centering
    \includegraphics[width=0.47\textwidth]{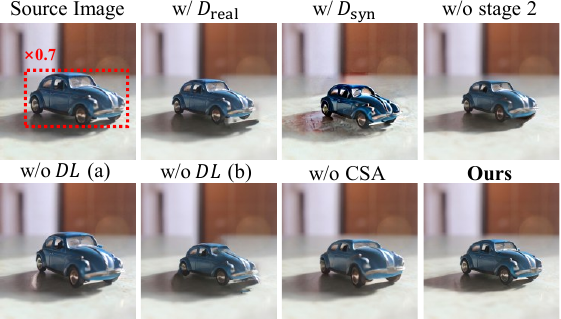}
    \caption{Ablation studies.}
    \label{fig:exp4}
\end{figure}

In this experiment, we perform ablation studies to illustrate the impacts of our multi-stage training strategy and \objectenhancercompletename (\objectenhancername). \cref{fig:exp4} compares different settings in the same editing scenario. Due to the unrealistic physical simulation, the model trained with $D_\text{real}$ manipulates the object in a copy-and-paste manner with incorrect shadow in the background. The model trained with $D_\text{syn}$ generates low-quality images with oversaturated color, which means it overfits to the rendering style. The performance of the model trained with a single stage falls between the above two models, resulting in unrealistic shadows. To verify the effectiveness of \domainadaptercompletename $DL_\text{syn}$ and $DL_\text{real}$, we additionally train a model based on a single set of LoRA modules. When the LoRA set is removed during inference ("w/o $DL$ (a)" in \cref{fig:exp4}), the model fails to perform the correct operation since the LoRA modules are coupled with operation modules. When the LoRA set is fully loaded ("w/o $DL$ (b)"), the image quality is compromised by the artifacts. On the other hand, the model without \objectenhancername exhibits lower consistency of object appearance. The object-level metrics in \cref{tab:ablation_study} also demonstrate the improvement in consistency from \objectenhancername.

\section{Conclusion}
We introduced \methodname, a 3D-aware autoregressive image editing framework that enables users to perform iterative object manipulations directly on real-world images. Trained on our dataset \datasetname that combines realistic and synthetic sequences across diverse objects and scenes, \methodname learns to perceive structural changes and maintain consistency across multiple editing rounds. By integrating condition-specific components and a multi-stage training scheme, our model can handle various 3D operations and generalize well to in-the-wild scenarios. Extensive experiments show that \methodname outperforms previous methods in both single-round and multi-round editing, delivering high-quality results.

\renewcommand{\thesection}{\Alph{section}}
\renewcommand{\thetable}{\Roman{table}}
\renewcommand{\thefigure}{\Roman{figure}}
\renewcommand{\theequation}{\roman{equation}}
\appendix
\section{Technical Appendices and Supplementary Material}\label{sec:appendix}

\subsection{Related works}
\label{sec:supp_related_works}
\noindent\textbf{Image Editing.}
Image editing aims to modify visual content within a source image. Leveraging the advances in text-to-image (T2I) generation~\cite{Imagen,StableDiffusion}, a series of diffusion-based methods~\cite{SEGA,T2I-Adapter,Localize-Object-Shape,SDEdit,MasaCtrl,DragDiffusion,DragonDiffusion,ding2025multimodal,he2025survey,shuai2024survey} have extended the pretrained models with multi-modal editing capabilities. For example, methods such as Prompt-to-Prompt~\cite{P2P}, InstructPix2Pix~\cite{InstructPix2Pix}, and LEDITS~\cite{Ledits} enable fine-grained modifications to subject appearance or image style while preserving the original layout. Other approaches, such as MagicRemover~\cite{MagicRemover}, focus on object removal. In addition, several works explore image inpainting and composition using T2I models, where new visual elements are injected into specific regions guided by text prompts~\cite{brushnet,Blended-Diffusion,RePaint,Differential-Diffusion,Blended-Latent-Diffusion,High-Resolution-Image-Editing} or reference images~\cite{Anydoor,Reference-Based-Composition,Paint-by-Example}.

\noindent\textbf{Object-centric Spatial Manipulation.}
Since text prompts are often insufficient to specify precise transformations, a line of research~\cite{CustomDiffusion360,Magic-Fixup} explores spatial cues for object-centric manipulation. Among them, dragging-based methods~\cite{FreeDrag,DragDiffusion,drag-your-noise} use source–target point pairs to animate local regions through non-rigid transformations. However, such an interface is limited in the ability to express object-level transformations and typically requires multiple point correspondences to indicate specific operations. To address these limitations, some approaches~\cite{DragonDiffusion,DiffEditor,DesignEdit,Diffusion-Self-Guidance} incorporate 2D masks to localize the objects before and after editing. Nevertheless, these methods are constrained to the image space and cannot handle operations that require 3D priors, such as object rotation. Other works~\cite{3DitScene,Image-Sculpting,DiffusionHandles,Diff3DEdit} attempt to reconstruct the scene structure, \eg, 3D Gaussian Splatting~\cite{3DGS} or point clouds, from a single image to enable arbitrary 3D-aware operations. However, these methods often suffer from time-consuming optimization~\cite{NeRF,3DGS} and noisy geometry estimation~\cite{StableDiffusion,Zero-1-to-3,SDS}, which degrade their practicality and output quality. Moreover, most of the above approaches struggle to produce realistic environmental interactions, such as shadows and occlusion relationships. Some recent studies leverage video or synthetic datasets to model such effects, but they often face difficulties in generalizing to real-world images~\cite{Object-3DIT,Zero-1-to-3} or supporting a wide variety of 3D operations~\cite{Object-3DIT,NeuralAssets}. Furthermore, existing methods exhibit inconsistencies across sequential edits because of their lack of scene structure awareness. This limitation significantly hinders their applicability in iterative editing scenarios.

\noindent\textbf{Video Generation Models.}
Video generation methods~\cite{LTX,CogVideoX,VideoCrafter1,VideoCrafter2,FFMC,AnyI2V} aim to create temporally coherent frames while preserving the desired visual content. Many recent approaches~\cite{AnimateDiff,Imagen-Video,VDM,Make-A-Video,SVD} incorporate temporal modules into network architectures to ensure cross-frame consistency, effectively capturing both foreground and background dynamics induced by camera or object motion. To improve training efficiency and realism in our editing framework, we leverage motion priors from the pretrained video diffusion model~\cite{SVD}, enabling our model to generate plausible environmental interactions during 3D-aware manipulation, such as changes in shadows and occlusions.

\begin{figure*}[h]
    \centering
    \includegraphics[width=0.999\linewidth]{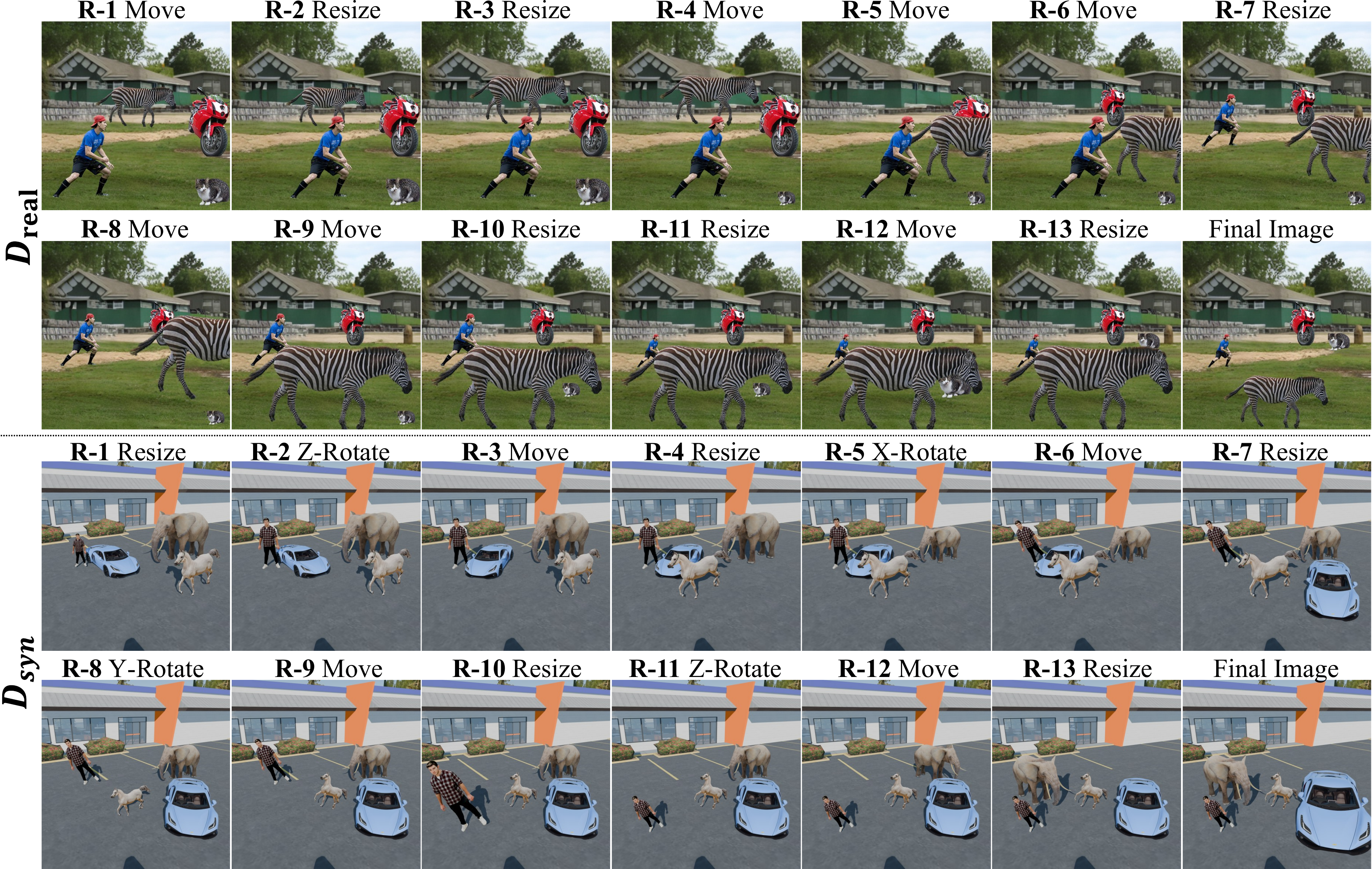}
    \caption{Examples of real and synthetic image sequences in the proposed \datasetname dataset.}
    \label{fig:dataset_pipeline}
\end{figure*}
\begin{table*}

  \setlength\tabcolsep{1mm}
  \begin{tabular}{lccccc}
    
    \specialrule{0.7pt}{0pt}{0pt}
    Method & Condition & Operation Types & Realistic Interaction & Multi-round & Reconstruction-free\\
    \hline
    \rowcolor{mygray2!70}Image Sculpting& \makecell[c]{2D mask + \\ \textbf{Transformation matrix}} & \textbf{All 3D operations} & \xmarkg & \xmarkg & \xmarkg \\
    3DitScene & \makecell[c]{2D mask + \\ \textbf{Transformation matrix}} & \textbf{All 3D operations} & \xmarkg & \xmarkg & \xmarkg\\
   \rowcolor{mygray2!70} Diff3DEdit& \makecell[c]{2D mask + \\ \textbf{Transformation matrix}} & \textbf{All 3D operations} & \xmarkg & \xmarkg & \xmarkg\\
    Diffusion Handles& \makecell[c]{2D mask + \\ \textbf{Transformation matrix}} & \textbf{All 3D operations} & \xmarkg & \xmarkg & \xmarkg\\
     \rowcolor{mygray2!70}Zero-1-to-3& \makecell[c]{2D mask + \\ \textbf{Transformation matrix}} & \makecell[c]{Translation, scaling, \\ $y/z$-axis rotation} & \xmarkg & \xmarkg & \cmark \\
     3DIT& Text prompt & \makecell[c]{Translation \\ $z$-axis rotation} & \cmark & \xmarkg & \cmark\\
    \rowcolor{mygray2!70}Neural Assets& \makecell[c]{3D bbox + \\ \textbf{Transformation matrix}} & \textbf{All 3D operations} & \cmark & \xmarkg & \cmark \\
    \textbf{\methodname(Ours)}& \makecell[c]{\textbf{2D bbox} + \\ \textbf{operation type\&value}} & \textbf{All 3D operations} & \cmark & \cmark & \cmark \\

    \hline
    \specialrule{0.7pt}{0pt}{0pt}
  \end{tabular}
  \vspace{-2mm}
  \caption{Summary of related methods in 3D-aware object manipulation. The compared methods are Image Sculpting~\cite{Image-Sculpting}, 3DitScene~\cite{3DitScene}, Diff3DEdit~\cite{Diff3DEdit}, Diffusion Handles~\cite{DiffusionHandles}, Zero-1-to-3~\cite{Zero-1-to-3}, 3DIT~\cite{Object-3DIT}, and Neural Assets~\cite{NeuralAssets}.}
  \label{tab:methods_summary}
\end{table*}
\subsection{Details of the Dataset}
\label{sec:dataset_details}
There is currently no well-established public dataset tailored for learning 3D-aware object manipulation in multi-round editing. To effectively support this task, a suitable dataset should satisfy the following key criteria:
\textbf{1) Diverse scene elements:} wide coverage of objects and backgrounds with variations in category, pose, and quantity to ensure generalization.
\textbf{2) Rich 3D operations:} accurate annotations for various 3D transformations, such as translation, scaling, and axis-aligned rotations.
\textbf{3) Sequential manipulations:} image sequences derived from consecutive 3D operations to support iterative editing.
\textbf{4) Realistic environmental interaction:} physically plausible changes (\eg, shadow, lighting, reflection).
{\textbf{5) Operation-exclusive scene dynamics:} the change of scene structure caused solely by object manipulations}.

A simple strategy for building 3D-aware manipulation data is to leverage existing video datasets~\cite{WebVid-10M,RealEstate-10K} and extract object transformations using off-the-shelf tools~\cite{RAFT,sam}. However, in-the-wild videos~\cite{MOSEv1,MeViSv1,MOSEv2,MeViSv2} often contain unwanted dynamics from the camera motion or non-rigid object deformation, violating criterion~\textbf{5}. Moreover, operations like rotation are poorly estimated, failing criterion~\textbf{2}. Synthetic data offers a controllable alternative. Object-3DIT~\cite{Object-3DIT} uses a 3D engine to generate labeled pairs but is limited by scene diversity, operation types, and lack of sequential edits, falling short on criteria~\textbf{1, 2,} and \textbf{3}. Additionally, relying solely on synthetic data harms generalization.

\begin{figure*}[h]
    \centering
    \includegraphics[width=0.999\linewidth]{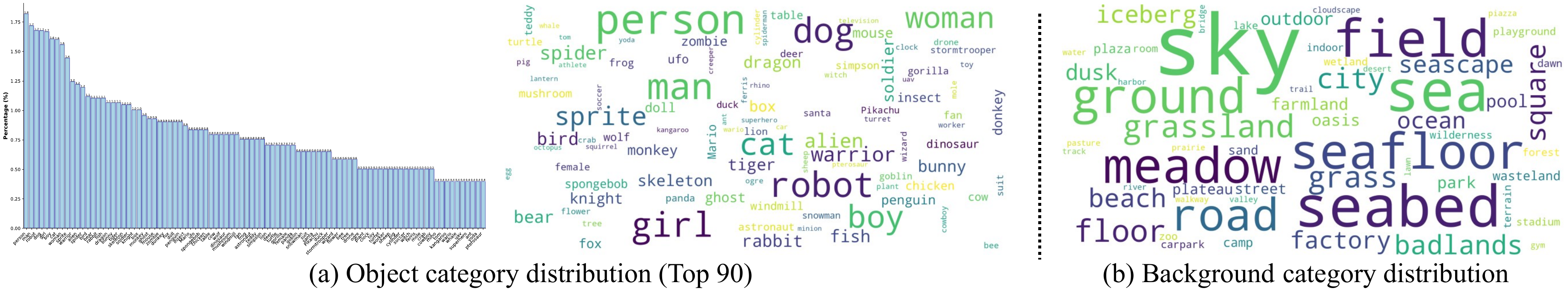}
    \caption{Statistics of $D_\text{syn}$ in \datasetname. (a) and (b) indicate the category distributions of objects and background, respectively.}
    \label{fig:statistic}
\end{figure*}

To address these issues, we introduce a hybrid dataset, \datasetname, which employs a rule-based algorithm to create image sequences induced by various 3D operations. Concretely, it collects scene elements from both realistic and synthetic domains, denoted as $D_\text{real}$ and $D_\text{syn}$, respectively. Each domain follows a three-step simulation process: \textit{Asset Collection}, where domain-specific objects and backgrounds are gathered; \textit{Scene Construction}, where an initial state is sampled; and \textit{Sequence Construction}, where operations from $O$ are iteratively applied, with state transitions governed by $p_{tf}$. Final observations are rendered using rendering functions $f_{m}$. It's worth noting that $O$, $p_{tf}$, and $f_m$ are specified by the domain, as illustrated in the "Data Generation" section.~\cref{fig:statistic} indicates the diversity of object and background categories in the synthetic part of \datasetname. Examples of the generated image sequences are shown in \cref{fig:dataset_pipeline}. We get 40K and 46K 32-length image sequences in $D_\text{real}$ and $D_\text{syn}$.

For $D_\text{real}$, we only consider translation $o^T$ and scaling $o^S$ as available operations. Given the shortest length $l$ of the background image sides, the center offsets along the width and height directions are all restricted in $[-0.6l,0.6l]$ for each step. The $d_\text{min}$ and $d_\text{max}$ are set to $10$ and $200$, while the depth offset is restricted in $[-30,30]$. The scaling factor $f_s$ is bounded within $[0.2,4]$, while the predefined minimal and maximal scales of each object $\hat{s}_\text{min}$ and $\hat{s}_\text{max}$ are set to $l/20$ and $l/4$, respectively. 

In $D_\text{syn}$, any operation from $\left\{o^T,o^S,o^X,o^Y,o^Z\right\}$ can be taken in each edit. For scene construction, we adjust the size of each object appropriately and randomly place them on the ground. For translation, objects are constrained to move along the ground surface without collision with each other, while maintaining visibility within the camera's view frustum. For scaling, we place the origin of the coordinate system at the bottom of the object to make the object remain grounded. The scaling factor in each step is bounded in $[0.2,4]$, and the rotation angles are restricted in $[-50,50]/[-45,45]/[-60,60]$ for $x/y/z$ axes. It's worth noting that the operation values will be resampled until no collision occurs in each step.

\begin{figure*}[t]
    \centering
    \includegraphics[width=\linewidth]{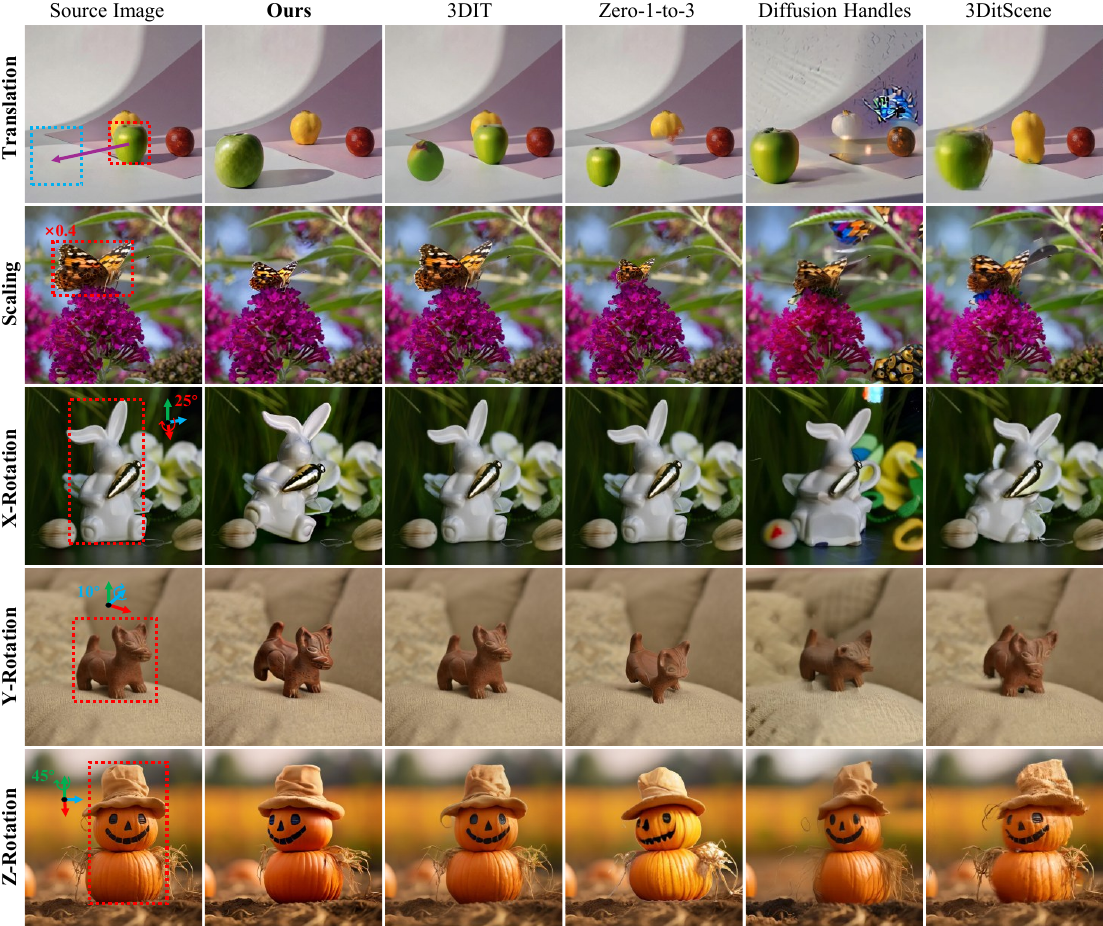}
    \caption{Evaluation of object effects under different 3D operations in single-round editing.}
    \label{fig:supp_exp1_1}
\end{figure*}

\begin{figure*}[t]
    \centering
    \includegraphics[width=\linewidth]{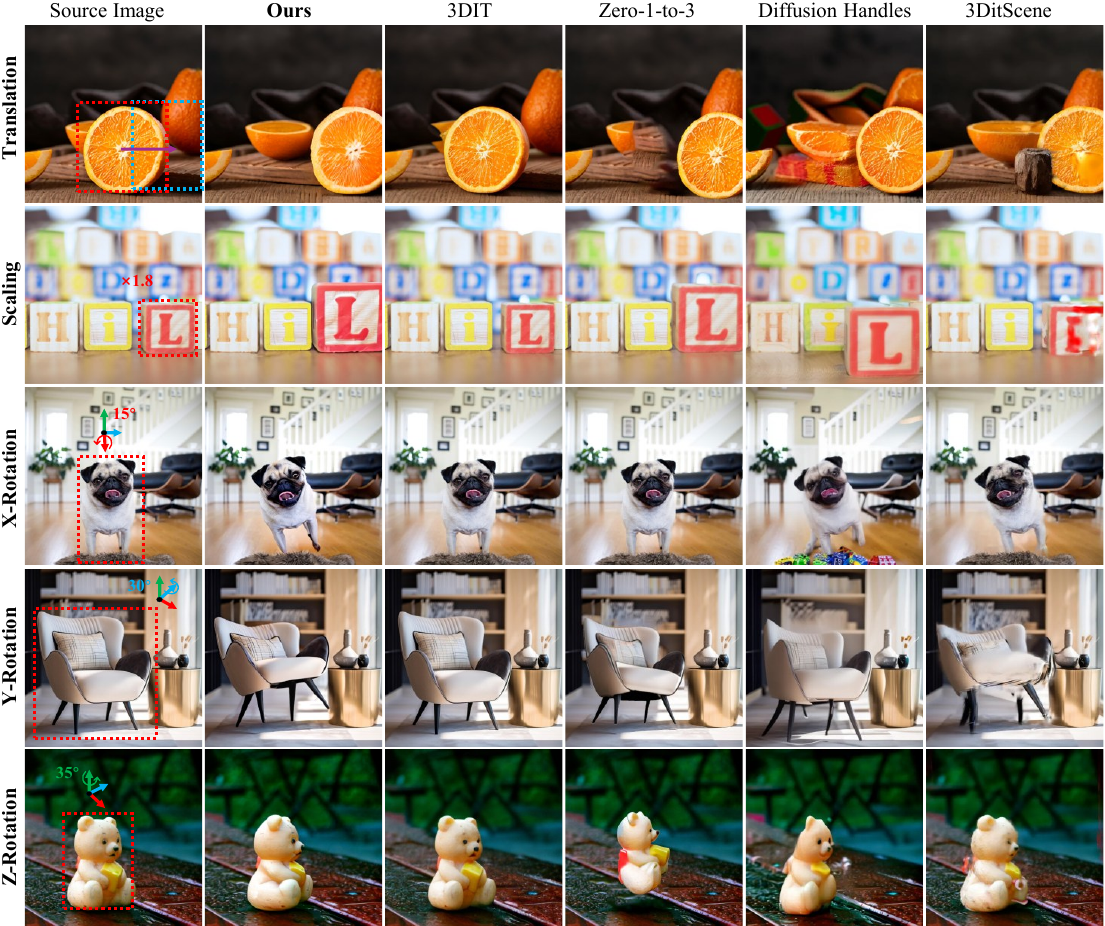}
    \caption{Evaluation of object effects under different 3D operations in single-round editing.}
    \label{fig:supp_exp1_2}
\end{figure*}

\subsection{Details of the Method}
\label{sec:method_details}

\noindent
\textbf{Differences with Related Methods.} \cref{tab:methods_summary} shows the differences of \methodname from related methods. We discuss Zero-1-to-3~\cite{Zero-1-to-3}, 3DIT~\cite{Object-3DIT} and Neural Assets~\cite{NeuralAssets} for \textit{image space} methods, while \textit{3D space} methods include Diffusion Handles~\cite{DiffusionHandles}, Diff3DEdit~\cite{Diff3DEdit}, Image Sculpting\cite{Image-Sculpting},and 3DitScene~\cite{3DitScene}.

Zero-1-to-3~\cite{Zero-1-to-3} estimates the novel view of the object through training on image pairs of the object captured from different cameras. However, the cameras located in a spherical coordinate system only differ in location, polar and azimuth angles, which hinders it to perform $x$-axis rotation. Besides, Zero-1-to-3 requires a tedious workflow to manipulate objects from real-world images. It has to perform an operation on the object region, then place the object into the inpainted background. Moreover, the image quality highly depends on the results from intermediate steps, while it can not create realistic background effects since there are no reliable techniques for object removal or inpainting.

3DIT~\cite{Object-3DIT} extends Zero-1-to-3 to achieve 3D-aware editing within an end-to-end framework. However, it only considers limited operation types like translation and $z$-axis rotation. Furthermore, the method receives a text prompt as the editing signal, which is inefficient to represent a precise manipulation and the source location of the object to be edited, impeding its application.

The above methods are all trained with synthetic datasets, leading to poor performance in real-world images. To alleviate the challenge, Neural Assets~\cite{NeuralAssets} was trained with a hybrid dataset containing both realistic and synthetic samples from public sources~\cite{Object-3DIT,objectron}. It encodes the appearance and pose of each object and background, and reconstructs the image through conditioning on these embeddings. For inference, it edits the image by injecting the new pose embedding. Nevertheless, the 3D bounding boxes are required in both training and inference processes, which are difficult to estimate for general objects and unfriendly for user input. Besides, the training data are constrained to limited categories and scene variations, which causes poor generalization ability.

In contrast to image space methods, which require to learn 3D priors from a designed dataset with comprehensive 3D annotations, a group of methods~\cite{Diff3DEdit,3DitScene,DiffusionHandles,Image-Sculpting} reconstructs the scene structure, which is flexible for various 3D operations. Most of these methods comply with the editing pipeline that consists of the following steps.

\textbf{1) 3D reconstruction.} This step obtains the scene structure from a single image. For example, Image Sculpting~\cite{Image-Sculpting} constructs NeRF~\cite{NeRF} through viewpoint-conditioned generation method~\cite{Zero-1-to-3} and Score Distillation Sampling (SDS)~\cite{SDS}. 3DitScene~\cite{3DitScene} iteratively estimates the novel view of the scene, and optimizes the 3DGS~\cite{3DGS} through reconstruction loss and SDS. Diff3DEdit~\cite{Diff3DEdit} and Diffusion Handles~\cite{DiffusionHandles} lift the image to 3D space through estimated depth from off-the-shelf tools~\cite{ZoeDepth}. 

\textbf{2) Object manipulation.} This step aims to operate the object for editing purposes. Image Sculpting translates the NeRF to meshes and performs deformation through binding the skeleton and transforming the bones. Differently, 3DitScene directly manipulates the 3D Gaussians belonging to the target object, while Diff3DEdit and Diffusion Handles operate on meshes and point clouds, respectively. 

\textbf{3) Image composition.} This step creates the final image by placing the transformed object in the target location seamlessly. Image Sculpting leverages personalization~\cite{DreamBooth} and feature injection techniques~\cite{PnP} to preserve the appearance and structure from the deformed object, while avoiding to combine the object and background in a copy-and-paste manner through feature blending. 3DitScene directly renders the edited scene from the camera view. For holes after depth-based warping, Diff3DEdit introduces a layered inpainting process to get a compatible image, while Diffusion Handles relies on the inherent geometry priors from a pretrained depth-conditioned model to imagine the ambiguous areas. Similarly, both of them introduce optimization processes~\cite{NTI,DreamBooth} to enhance consistency.

Although 3D space methods can achieve a wide range of 3D-aware operations effectively, they are constrained by cumbersome workflows and time-consuming optimization processes. Moreover, the image quality is highly affected by the performance of estimation tools used in intermediate steps~\cite{SDS,DreamBooth,SDEdit,NTI,Zero-1-to-3,ZoeDepth,NeRF,StableDiffusion}, while the noisy estimations may lead to artifacts and accumulated errors in outcomes. Additionally, most of these methods struggle to generate desirable background effects due to the lack of physical priors.

Moreover, existing methods from both algorithm families can not handle multi-round image editing since they are unaware of scene structure changes. For example, 3DIT was trained with image pairs within a single round, while other 3D space methods reconstruct the scene without previous structure information in each edit. In contrast, our \methodname outperforms in generating precise object/background effects and maintaining consistency after multiple edits. Besides, \methodname offers a friendly user interface to present the operation and source region precisely, and manipulates the object without additional optimization processes.

\begin{figure*}
    \centering
    \includegraphics[width=\linewidth]{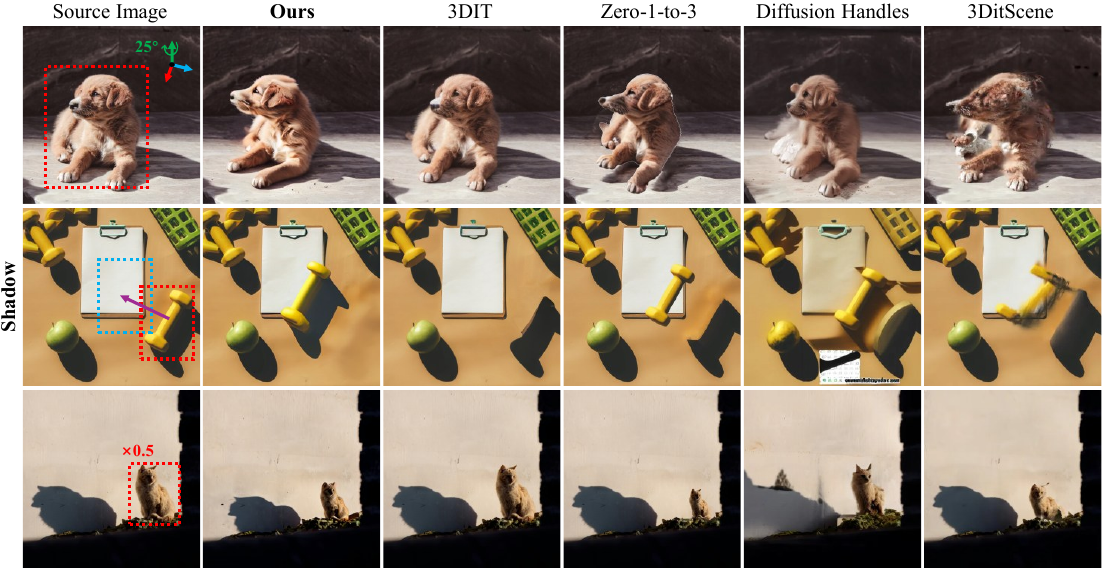}
    \caption{Evaluation of shadow effects in single-round editing.}
    \label{fig:supp_exp2_1}
\end{figure*}

\begin{figure*}
    \centering
    \includegraphics[width=\linewidth]{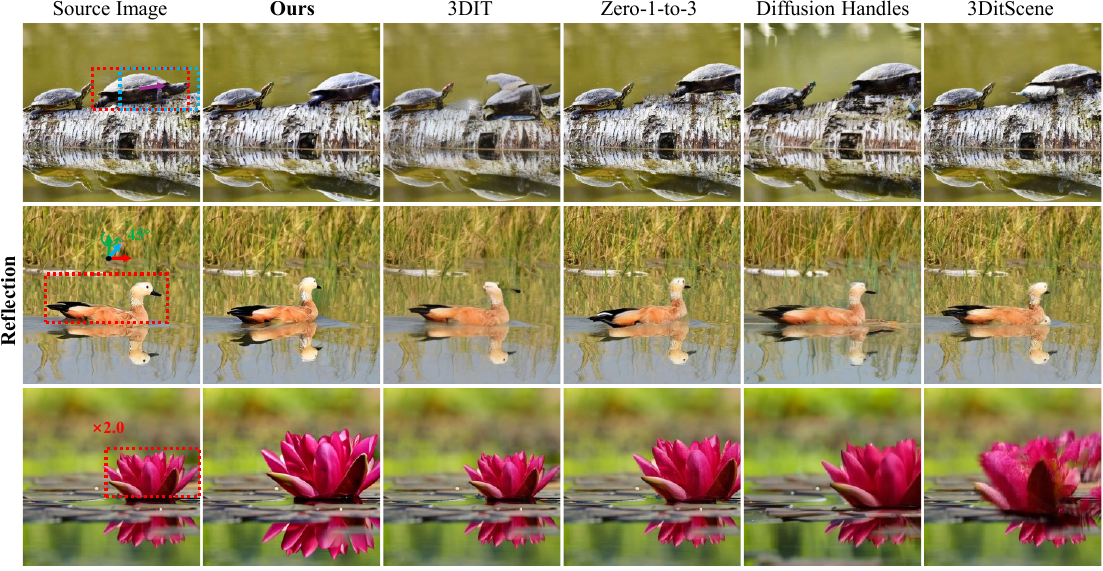}
    \caption{Evaluation of reflection effects in single-round editing.}
    \label{fig:supp_exp2_2}
\end{figure*}

\begin{figure*}
    \centering
    \includegraphics[width=\linewidth]{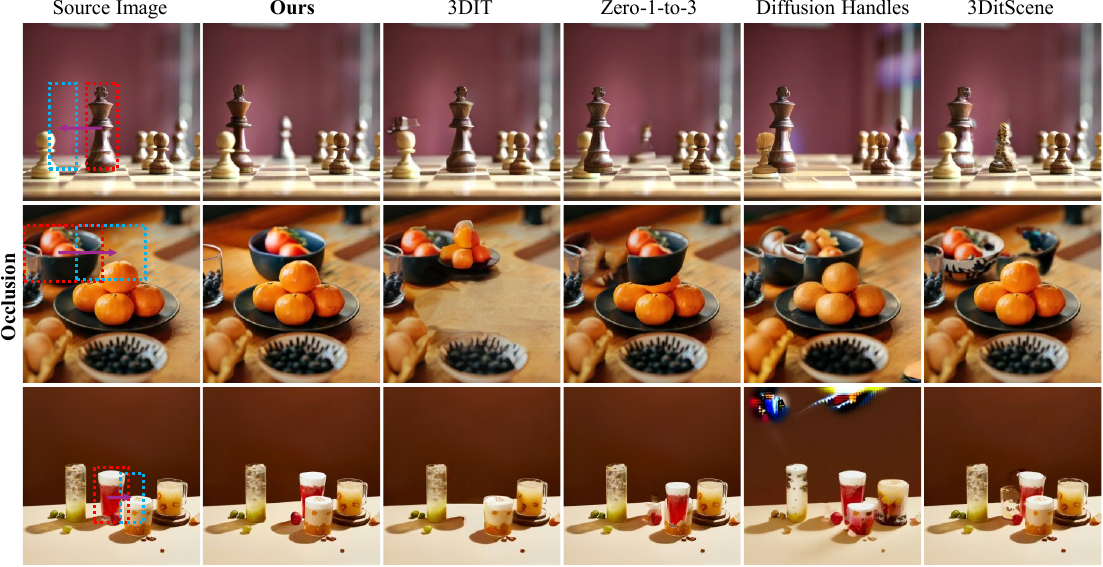}
    \caption{Evaluation of occlusion effects in single-round editing.}
    \label{fig:supp_exp2_3}
\end{figure*}

\begin{figure}[t]
    \centering
    \includegraphics[width=0.999\linewidth]{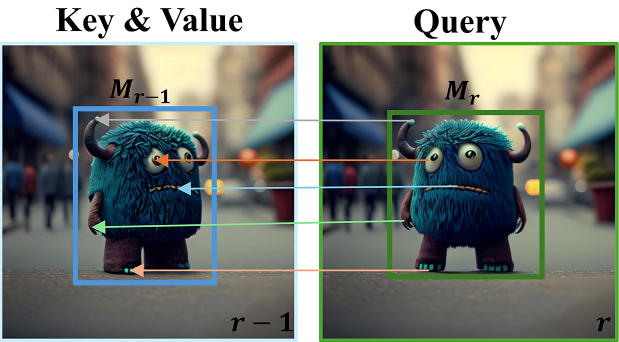}
    \caption{Illustration of \objectenhancercompletename.}
    \label{fig:enhancer}
\end{figure}

\noindent
\textbf{Model Architecture.} We introduce the following modules into the base model~\cite{SVD}. First, the \frameencodercompletename is a lightweight network that consists of a stack of residual blocks, processing the current timestep, along with the concatenation of previous observations and the target region in the current edit. The output from \frameencodercompletename is only added to the first layer from down blocks. The \operationencodercompletename projects the \textit{source region} and \textit{operation type\&value} through fourier embedding and MLP, where the output dimension of the two-layer MLP is set to 768. Specifically, the coordinates of bounding boxes and centers are normalized in $[0,1]$, and our method can handle negative coordinate values to address cases where parts of the object are outside the field of view. Before feeding the fourier embedding into MLP, we sum the condition $c$ (\eg $l_i^p$ or $o_i^T$) of each iteration and corresponding ``null'' embedding $e_{null}$ through $m\text{Fourier}(c)+(1-m)e_{null}$, where $m=1$ indicates that the condition is given in that iteration. All conditions of the first frame are set to ``null'' since it has no reference frame. Finally, the outputs from all conditions are concatenated along the channel dimension, and the features of each iteration are then applied sequence-wise concatenation. To inject operation features into the base model, \operationselfattentioncompeletename~follows \cite{GLIGEN} and performs self-attention calculation as in Eq.2. $\gamma$ is initialized as 0 and $\beta$ is set to $1$ during training. Furthermore, we enhance the ordinary self-attention layers by \objectenhancercompletename, which constructs correspondence between the same objects from the current and the last frames. $\lambda$ from Eq.3 is also initialized as 0. \figurename~\ref{fig:enhancer} illustrates our attention design.

\noindent
\textbf{Inference Details.}
Specifically, in the $i$-th editing round, a new user request is appended to the operation buffer, while the corresponding historical context $h_i$ (comprising previously generated observations and user operations) is retrieved from both the frame and operation buffers. \methodname then processes this context $h_i$ to generate the updated observation $x_i$. Although intermediate outputs $x_{0:i-1}$ are also produced, only the current frame $x_i$ is retained in the frame buffer, in order to reduce the accumulation of reconstruction errors over multiple rounds. To further improve efficiency in long-term interaction scenarios, users can choose the previous few historical frames and operations from the buffers as conditional input, thereby reducing computational overhead. It is worth noting that \domainadaptercompletename modules are omitted during inference to preserve the quality of the base generative model. The inference pipeline of our method is illustrated in \textbf{Algorithm}~\ref{alg:inference}.

\begin{algorithm}[t]
\caption{Inference pipeline of \methodname.}
\label{alg:inference}
\textbf{Input:} Source image $x_0$; maximal buffer size $N$;\\
\textbf{Output:} The frame buffer $\mathcal{B}_{f}$.  
\begin{algorithmic}[1] 
\STATE Initialize the frame buffer $\mathcal{B}_f=[x_0]$ and operation buffer $\mathcal{B}_o=[\ \ ]$;\\
\STATE $r=0$;\\
\WHILE{$\text{True}$}
\STATE Obtain the current operation $o_r$ from user;\\
\STATE Append the $o_r$ into $\mathcal{B}_o$;\\
\STATE $r=r+1$;\\
\STATE Construct the history $h_r$ from $\mathcal{B}_f$ and $\mathcal{B}_o$; \\
\IF{$\textrm{length}(h_r)>N$}{
    \STATE Discard the first $(\text{length}(\mathcal{B}_f)-N)$ observation-operation pairs $\left\{x_i,o_i\right\}_{i=0}^{\text{length}(\mathcal{B}_f)-N-1}$ from $h_r$;\\
}
\ENDIF
\STATE Get the initial noise $\epsilon$ sampled from $\mathcal{N}(\mathbf{0},\mathbf{I})$;\\
\STATE Obtain the estimated observations $x_{0:r}$ through diffusion sampling based on $\epsilon,h_r$;\\
\STATE Extract the current observation $x_r$ from $x_{0:r}$ and append it to $\mathcal{B}_f$;\\
\ENDWHILE
\STATE \textbf{return} $\mathcal{B}_{f}$

\end{algorithmic}
\end{algorithm}

\begin{figure*}[htbp]
    \centering
    \includegraphics[width=0.999\linewidth]{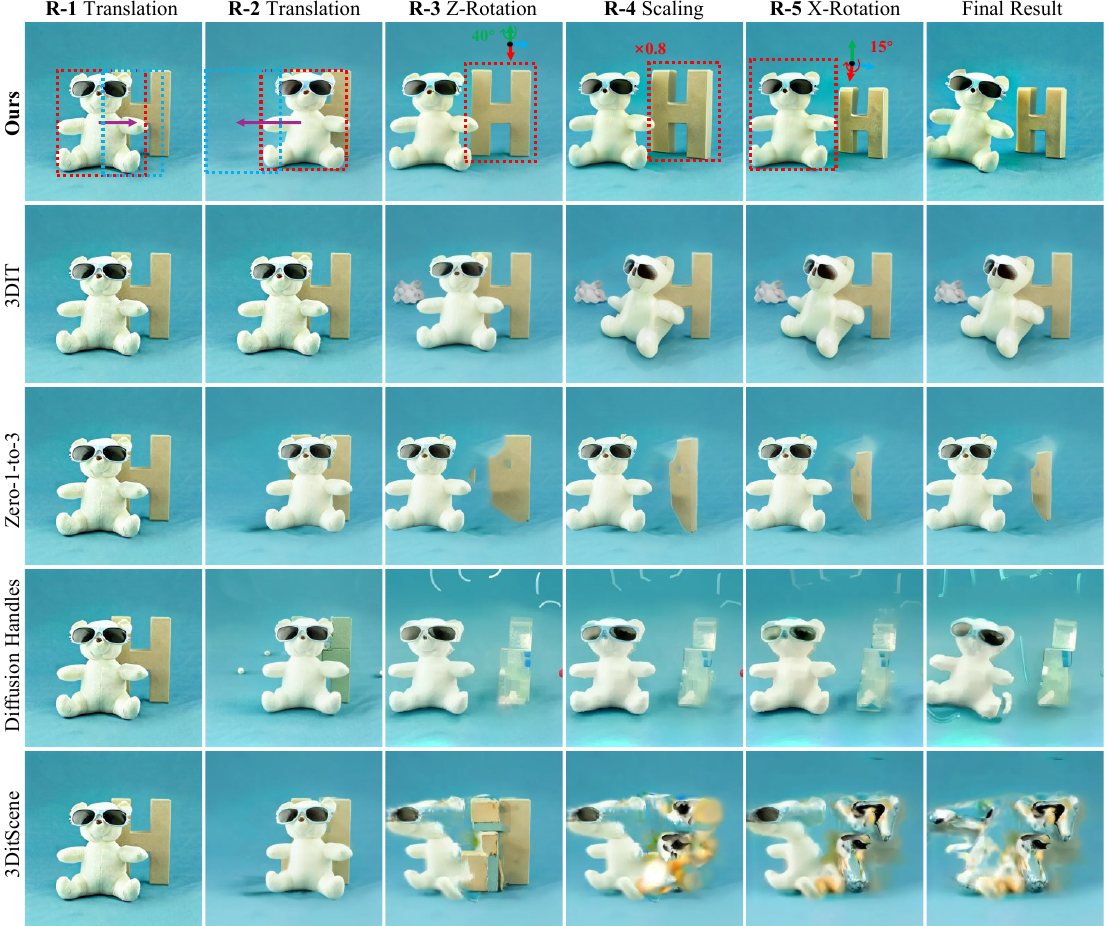}
    \caption{Evaluation in multi-round editing.}
    \label{fig:supp_exp3}
\end{figure*}

\begin{figure*}[htbp]
    \centering
    \includegraphics[width=0.999\linewidth]{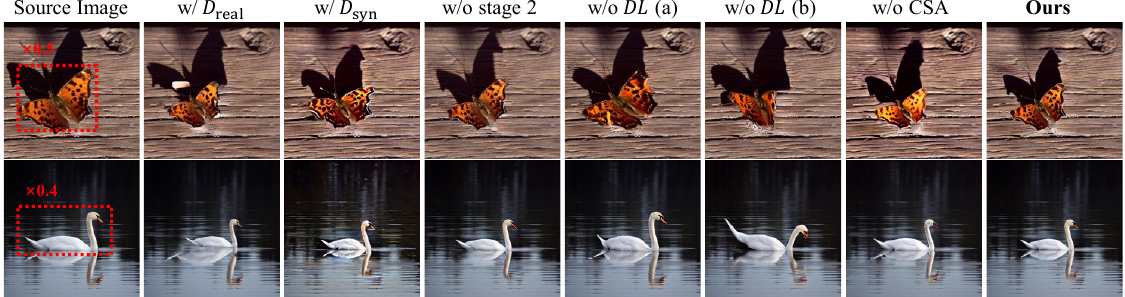}
    \caption{Ablation studies.}
    \label{fig:supp_exp4}
\end{figure*}

\subsection{More Comparisons and Analyses}
\label{sec:more_comparison}

\noindent
\textbf{Details about Validation Data.} All images in validation are sourced from Unsplash~\cite{Unsplash}, Pixabay~\cite{pixabay}, and Pexels~\cite{Pexels} under their respective free-use licenses, which permit non-commercial and commercial use without attribution.

\noindent
\textbf{User Study.} We conduct a user study to assess human preferences in the following aspects. \textbf{1) Image quality}. The edited image has to maintain the same quality as the source image. \textbf{2) Operation fidelity.} The object and background effects caused by the operation need to comply with the physical laws in the real world. \textbf{3) Consistency.} Each scene element has to remain consistent across edits. \tablename~\ref{tab:user_study} shows the results of the user study. Specifically, we employed 30 volunteers to evaluate outcomes from each method. For each score, we average the results and normalize the value by the maximum value.

\begin{table}[t]
\centering

\setlength\tabcolsep{1mm}
\begin{tabular}{l|c|c|c|c|c}
\specialrule{0.7pt}{0pt}{0pt}
Metric & 3DIT & Zero-1-to-3  & DH & 3DS & \textbf{\methodname} \\
\hline
{Image quality} & 0.75 & \underline{0.88} & 0.81 & 0.85 & \textbf{0.93} \\
{Object effects} & 0.44 & 0.79 & 0.88 & \underline{0.90} & \textbf{0.94} \\
{Background effects} & \underline{0.59} & 0.03 & 0.33 & 0.11 & \textbf{0.98} \\
{Scene consistency} & 0.32 & \underline{0.35} & 0.12 & 0.09 & \textbf{0.91} \\
\specialrule{0.7pt}{0pt}{0pt}
\end{tabular}
\caption{Human preferences in quality, object effects, background effects, and scene consistency. The compared methods are 3DIT~\cite{Object-3DIT}, Zero-1-to-3~\cite{Zero-1-to-3}, Diffusion Handles (DH)~\cite{DiffusionHandles}, and 3DitScene (3DS)~\cite{3DitScene}.}
\label{tab:user_study}
\end{table}

\noindent
\textbf{More Qualitative Experiments.} ~\cref{fig:supp_exp1_1}-\cref{fig:supp_exp1_2} evaluate the performance of compared methods in terms of object effects. The results demonstrate that \methodname outperforms in processing various 3D operations. The compared methods are unable to obtain desirable outcomes under most of the manipulations. For translation, only \methodname recovers the occluded region, such as the yellow fruit in row 1 of \cref{fig:supp_exp1_1} and the orange in row 1 from \cref{fig:supp_exp1_2}. Although 3D space methods~\cite{3DitScene,DiffusionHandles} can handle rotations effectively, they suffer from time-consuming optimization processes and suboptimal image quality caused by noisy geometry estimation, as shown in columns 5-6 from rows 3-5 in \cref{fig:supp_exp1_1} and \cref{fig:supp_exp1_2}. For example, due to performing on incomplete point clouds, Diffusion Handles fails to generate structurally intact objects when handling large rotations. 3DitScene tends to generate edited objects with blurred edges.

Besides, \cref{fig:supp_exp2_1}-\cref{fig:supp_exp2_3} evaluate the background effects from generated images. Creating realistic environmental interaction is difficult for Zero-1-to-3~\cite{Zero-1-to-3}, since there are no reliable tools for removing/generating object effects in the source/target location. On the other hand, Diffusion Handles relies on limited physical priors from the base model, leading to suboptimal results. In contrast, only \methodname generates precise object and background effects under arbitrary operations.

Additionally, \cref{fig:supp_exp3} represents the methods' performance in multi-round image editing. As shown in the first 3 columns, all compared methods can not recover the letter ``H'' behind the bear. Besides, accumulated errors hamper the subsequent manipulations. In comparison, \methodname accomplishes the operation in each iteration, while exhibiting high consistency of scene elements across edits.

\subsection{More Ablation Studies}
\label{sec:more_ablation_study}
In addition, we provide more results of ablation studies in \cref{fig:supp_exp4}. Training solely with $D_\text{real}$ or $D_\text{syn}$ leads to suboptimal results in terms of reality or image quality. This is caused by insufficient physical simulation of each domain, like incorrect environmental interaction from $D_\text{real}$ and unrealistic rendering in $D_\text{syn}$. In contrast, training the model in the whole dataset with \domainadaptercompletename helps to learn the object effects shared among the domains, preventing the control modules to learn the domain-specific content style. As illustrated in columns 5-6 of \cref{fig:supp_exp4}, when training solely with a single set of LoRA, the method fails to generate high-fidelity and high-quality results. Furthermore, the second stage improves the performance in the generation of background effects through finetuning the model on $D_\text{syn}$. Finally, the experiment also shows the effectiveness of \objectenhancercompletename (\objectenhancername), which enhances the appearance consistency by associating the features of the edited object in the current and last frames.

\subsection{Limitations}\label{sec:limitation}
Although our method enables effective multi-round 3D-aware manipulation, it can not accomplish non-rigid deformations. Furthermore, the method is constrained by the number of editing iterations. Excessive editing steps lead to unacceptable memory consumption and computational overhead. While we mitigate this issue by only conditioning on partial historical information, it inevitably compromises consistency. Thus, we will explore how to get optimal trade-offs between efficiency and consistency in multi-round editing in our future work.

\newpage
\bibliography{aaai2026}

\end{document}